\documentclass{article} 
\usepackage{iclr2027_conference,times}

\usepackage{amsmath,amsfonts,bm}

\def\eqref#1{equation~\ref{#1}}

\def\1{\bm{1}}

\def\vz{{\bm{z}}}

\DeclareMathAlphabet{\mathsfit}{\encodingdefault}{\sfdefault}{m}{sl}
\SetMathAlphabet{\mathsfit}{bold}{\encodingdefault}{\sfdefault}{bx}{n}

\usepackage{hyperref}
\usepackage{url}

\hypersetup{hidelinks}
\usepackage{subcaption}
\usepackage{booktabs, multirow, makecell} 
\usepackage{xcolor} 
\usepackage{graphicx}
\usepackage{wrapfig} 
\usepackage{adjustbox} 
\usepackage{afterpage} 
\usepackage[capitalize,nameinlink]{cleveref}
\crefname{figure}{Fig.}{Figs.}
\Crefname{figure}{Fig.}{Figs.}
\crefname{table}{Tab.}{Tabs.}
\Crefname{table}{Tab.}{Tabs.}
\crefname{section}{Sec.}{Secs.}
\Crefname{section}{Sec.}{Secs.}
\crefname{subsection}{Sec.}{Secs.}
\Crefname{subsection}{Sec.}{Secs.}

\definecolor{poscolordiff}{RGB}{0,120,0}
\definecolor{negcolordiff}{RGB}{180,0,0}
\newcommand{\dpos}[1]{{\scriptsize(\textcolor{poscolordiff}{$#1$})}}
\newcommand{\dneg}[1]{{\scriptsize(\textcolor{negcolordiff}{$#1$})}}
\newcommand{\dnone}[1]{{\scriptsize($#1$)}}

\title{AdaptMIL: Domain Adaptation for Multiple Instance Learning in Digital Pathology}

\author{\normalfont
\textbf{Ben Vardi$^{1}$,
Dana Schonberger$^{2}$,
Yuval Friedmann$^{2}$,
Zohar Yakhini$^{1,3}$,}\\
\textbf{Iris Barshack$^{2,4}$,
Alexander Loebel$^{2}$,
Ariel Shamir$^{1}$} \\
$^{1}$Reichman University \\
$^{2}$Institute of Pathology, Sheba Medical Center \\
$^{3}$Technion -- Israel Institute of Technology \\
$^{4}$Ariel University \\
}

\newif\ifarxiv
\arxivtrue  

\ifarxiv
    \iclrfinalcopy
\fi

\begin{document}

\ifarxiv
    \raggedbottom
\fi
\maketitle
\ifarxiv
    \lhead{}
    \renewcommand{\headrulewidth}{0pt}
\fi

\begin{abstract}

Pathology foundation models (PFMs) support a wide range of histopathology tasks on whole-slide images through multiple instance learning (MIL).
Despite recent progress,
these models remain sensitive to cross-hospital distribution shifts:
performance often degrades when a task head (e.g., classification) is trained on data from one hospital and then used on data from another hospital.
To address this challenge, we propose AdaptMIL, a training strategy for integrating domain adaptation into MIL training.
AdaptMIL is motivated by our finding that directly applying domain adaptation objectives during MIL training can interfere with MIL aggregation learning,
often resulting in performance degradation over standard training.
AdaptMIL resolves this by decoupling representation-level domain alignment learning from MIL aggregation learning through a two-stage training procedure.
We show that AdaptMIL improves cross-hospital performance and outperforms alternative adaptation methods across experiments spanning multiple WSI datasets, tasks, PFMs, MIL architectures, and domain adaptation objectives. AdaptMIL's code will be released upon publication.

\end{abstract}

\section{Introduction}

The emergence of modern pathology foundation models (PFMs) has driven substantial progress in digital pathology in recent years~\citep{chen2024towards,lu2024visual,filiot2025distilling}.
These models take digitized tissue images as input and produce general-purpose representations that can support a wide range of downstream histopathology tasks, including cancer detection, tumor subtyping, biomarker prediction, and survival analysis~\citep{neidlinger2025benchmarking}.

Unlike conventional image encoders that process standard-sized images, PFMs are used to process extremely large images, commonly referred to as whole-slide images (WSIs), which can contain billions of pixels.
This requires a specialized processing pipeline: a WSI is first divided into smaller patches, each patch is independently encoded by a PFM into a patch-level representation, and finally, the patch representations are aggregated into a single slide-level embedding using a weakly supervised multiple instance learning (MIL) module, most commonly trained from scratch per task~\citep{wang2024advances,shao2025do}.

Despite their progress, PFM-based MIL pipelines may still struggle in real-world deployment conditions.
A major challenge is their limited robustness to variability across hospitals from which tissue samples were collected, arising from differences in tissue processing protocols~\citep{komen2025robustfoundationmodelsdigital}.
As a result, the performance of task heads (e.g., classifiers) built on PFM representations degrades when the MIL module and task head are trained on data from one hospital and used on data from another~\citep{fernandez2026domain,umer2026multicenterbenchmarkmultipleinstance}.
This poses a practical limitation, since labeled data from a specific \emph{target hospital} is often unavailable
due to costly and time-consuming expert annotation,
whereas unlabeled data from that hospital may be readily available.
Meanwhile, labeled data from other \emph{source hospitals} may already be available, a setting known as \emph{unsupervised domain adaptation} (UDA)~\citep{wilson2020survey}.

A direct approach is to fine-tune the pretrained PFM toward the target hospital domain.
This can be done using domain adaptation objectives in the UDA setting,
and requires labels for individual image patches from source-domain WSIs.
Although this approach has been explored before~\citep{vardi2026discrepancyminimizationimprovescrosshospital}, its applicability is limited by the scarcity of patch-level annotations in pathology.
By contrast, slide-level annotations are more readily available and can be used for end-to-end PFM fine-tuning. However, this requires propagating gradients from the slide-level objective through the MIL module and back through the PFM across hundreds or thousands of patches per slide, substantially increasing the computational cost compared with training the MIL module alone~\citep{campanella2025real,kumar2026single}.

\begin{wrapfigure}{r}{0.5\textwidth}
    \centering
    \includegraphics[width=\linewidth]{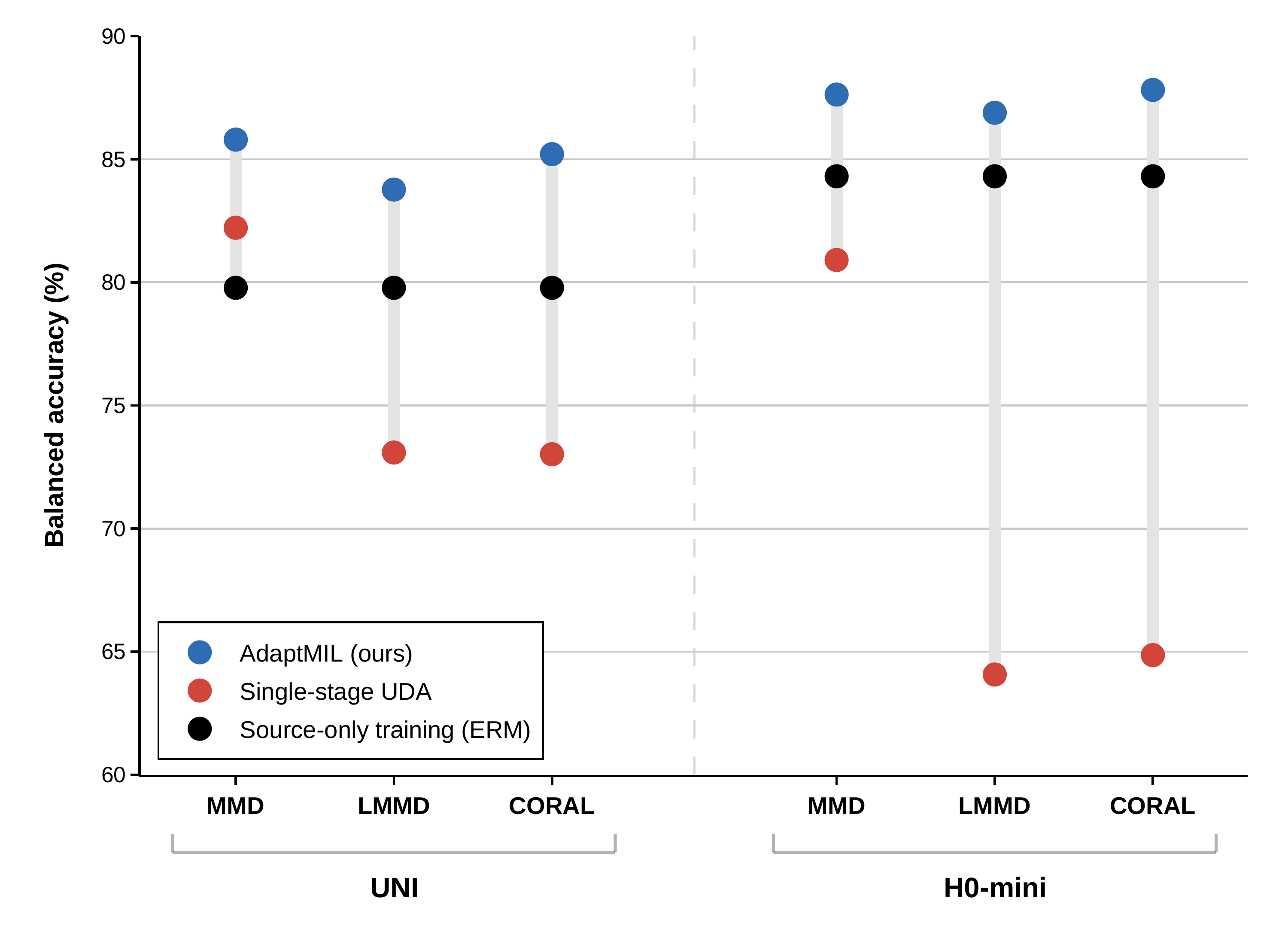}
    \caption{
    AdaptMIL improves cross-hospital performance across PFMs such as UNI and H0-mini and across domain adaptation objectives.
    It is motivated by the observation that directly applying standard UDA during MIL training often degrades performance.
    }
\label{fig:adaptation_results}
\end{wrapfigure}

In this work, we pursue a simpler alternative: we keep the pretrained PFM frozen and perform adaptation within the MIL training pipeline, using only slide-level supervision.
We find that a standard approach to unsupervised domain adaptation, which combines a task loss on the labeled source domain and a domain alignment objective to align the source and target domains, can fail when applied to MIL pipelines.
In particular, we find that domain alignment and learned MIL aggregation can interfere with one another: rather than learning more domain-invariant representations, the MIL module reduces the domain discrepancy by altering its aggregation pattern.
This can result in performance below source-only training.

To address this, we propose \textbf{AdaptMIL}, a two-stage MIL training strategy for domain adaptation that decouples domain alignment from aggregation learning.
The first stage trains a patch-level projection layer to make representations more domain-aligned and uses fixed mean pooling to aggregate them into slide-level representations.
Source--target domain alignment is performed on these slide-level representations together with a source task objective that preserves task-discriminative information.
In the second stage, the MIL aggregation is trained using the source task objective on top of the adapted projection layer learned in the first stage.
By performing domain alignment before learning the aggregation mechanism, AdaptMIL prevents the MIL module from altering its aggregation pattern to satisfy the alignment objective, 
resulting in gains over source-only training (\cref{fig:adaptation_results}).

Across experiments on multi-hospital datasets, we show that AdaptMIL outperforms source-only MIL training and alternative adaptation methods.
We demonstrate these gains across multiple PFMs, slide-level tasks, domain alignment objectives, and MIL aggregation methods.
Our code will be released upon publication.

Our contributions are:
\begin{itemize}
    \item We explore UDA in slide-level pathology MIL pipelines and identify a failure mode for jointly learning domain alignment and MIL aggregation.
    \item We introduce AdaptMIL, a two-stage training strategy for MIL that decouples domain alignment from aggregation learning and can be applied across different MIL architectures.
    \item We evaluate AdaptMIL across multiple pathology tasks, WSI datasets, PFMs, MIL architectures, and domain alignment objectives, and demonstrate improved cross-hospital performance over source-only training and other adaptation approaches.
\end{itemize}

\section{Related work}
\label{sec:related_work}

\subsection{Pathology foundation models and MIL}
\label{subsec:related_work:pfms_mil}

Pathology foundation models (PFMs) encode hematoxylin and eosin (H\&E)-stained histopathology images into general-purpose embeddings that support a wide range of downstream tasks~\citep{chen2024towards,lu2024visual,filiot2025distilling,vorontsov2024foundation}.
These representations are learned through large-scale pretraining, using objectives such as self-supervision~\citep{chen2024towards,vorontsov2024foundation} and vision-language contrastive learning~\citep{lu2024visual}.
In pathology practice, inference is performed on whole-slide images (WSIs), which often reach gigapixel scale (e.g., $100{,}000 \times 100{,}000$ pixels).
The predominant pipeline therefore segments the WSI, divides it into patches, encodes the patches independently using a PFM, and then aggregates the patch embeddings into a slide-level embedding using a multiple instance learning (MIL) module~\citep{wang2024advances}.

MIL addresses learning from bags of instances for which supervision is available only at the bag level~\citep{dietterich1997solving}.
In digital pathology, a WSI is treated as a bag of patches~\citep{campanella2019clinical}: while a WSI-level label may indicate, for example, the presence of cancer, labels for the individual patches are generally unavailable.
MIL methods therefore aggregate patch embeddings into a single slide-level representation from which the downstream task can be predicted.

A wide range of MIL methods have been proposed, varying along several dimensions, such as their aggregation mechanisms, architectures, and whether the aggregation is learnable or parameter-free.
An influential architecture is ABMIL~\citep{ilse2018abmil}, which learns a network that assigns attention weights to instances and aggregates them through a weighted average.
Subsequent methods introduce more sophisticated aggregation mechanisms.
For example, DSMIL~\citep{li2021dual} identifies a critical instance and uses it to guide attention over the remaining instances, while TransMIL~\citep{shao2021transmil} employs transformer-based attention to model dependencies between patches.
A common practice is to train MIL models from scratch for each downstream task, while more recent work has explored pretraining slide-level MIL models~\citep{ding2025multimodal, shao2025do}.

\subsection{Cross-hospital robustness}

Despite impressive performance, slide-level PFM pipelines may face limitations in real-world settings.
A major challenge emerges in multi-hospital settings due to variation in H\&E appearance, driven by differences in tissue processing, staining, and digitization protocols~\citep{howard2021impact,thiringer2026scannerinduceddomainshiftsundermine}.

Prior work has examined cross-hospital robustness at the patch level, showing that PFM embeddings encode hospital-specific signatures and exhibit related distribution shifts~\citep{komen2025robustfoundationmodelsdigital}.
Proposed approaches to mitigate these shifts include test-time stain normalization of target patches~\citep{macenko2009method,reinhard2002color}, training patch-level models with domain adaptation objectives~\citep{vray2024distill}, fine-tuning PFMs for domain alignment~\citep{vardi2026discrepancyminimizationimprovescrosshospital}, and learning transformations on top of frozen PFM features~\citep{komen2025robustfoundationmodelsdigital}.

Cross-hospital performance degradation has also been observed in slide-level tasks~\citep{fernandez2026domain,umer2026multicenterbenchmarkmultipleinstance}.
Several patch-level solutions can be transferred to slide-level prediction: stain normalization can be applied across all patches, while adapted PFMs can be used within MIL pipelines.
However, direct slide-level adaptation remains comparatively underexplored.
HASD~\citep{liu2025hasd} takes this direction and proposes a pathology-specific hierarchical adaptation framework with several specialized objectives.
We also perform adaptation at the slide-aggregation level, but rather than introducing a specialized adaptation objective, we propose a training strategy that enables established domain adaptation objectives to be integrated into MIL.

\subsection{Domain adaptation}
\label{subsec:related_work:da}

Domain adaptation~\citep{Wang_2018} aims to transfer a model trained on data from a source domain to a different target domain.
Unsupervised domain adaptation (UDA)~\citep{wilson2020survey} is a common setting, where labeled source data is available, while only unlabeled data is available in the target domain.

A central approach to UDA is domain-invariant feature learning, which aims to reduce differences between source and target representation distributions.
Invariant representations can be learned through discrepancy minimization methods, which minimize a differentiable distance metric between the two distributions.
Prominent examples include Maximum Mean Discrepancy (MMD)~\citep{long2015learning}, local MMD (LMMD)~\citep{zhu2020deep}, and CORAL~\citep{sun2016deep}.
Other domain-invariant methods use different mechanisms, such as domain-adversarial neural networks (DANN)~\citep{ganin2016domain}, which learns domain-invariant features by making representations from the source and target domains difficult to distinguish.

Although widely used across computer vision and machine learning, the interaction between domain adaptation and MIL remains less explored in slide-level pathology.
In this work, we investigate this interaction and develop a training strategy for domain adaptation within MIL pipelines.

\section{Method}
\subsection{Problem setup and domain adaptation for MIL}
\label{subsec:method:setup}

\begin{figure}[t]
    \centering
    \begin{subfigure}[t]{0.48\linewidth}
        \centering
        \includegraphics[width=\linewidth]{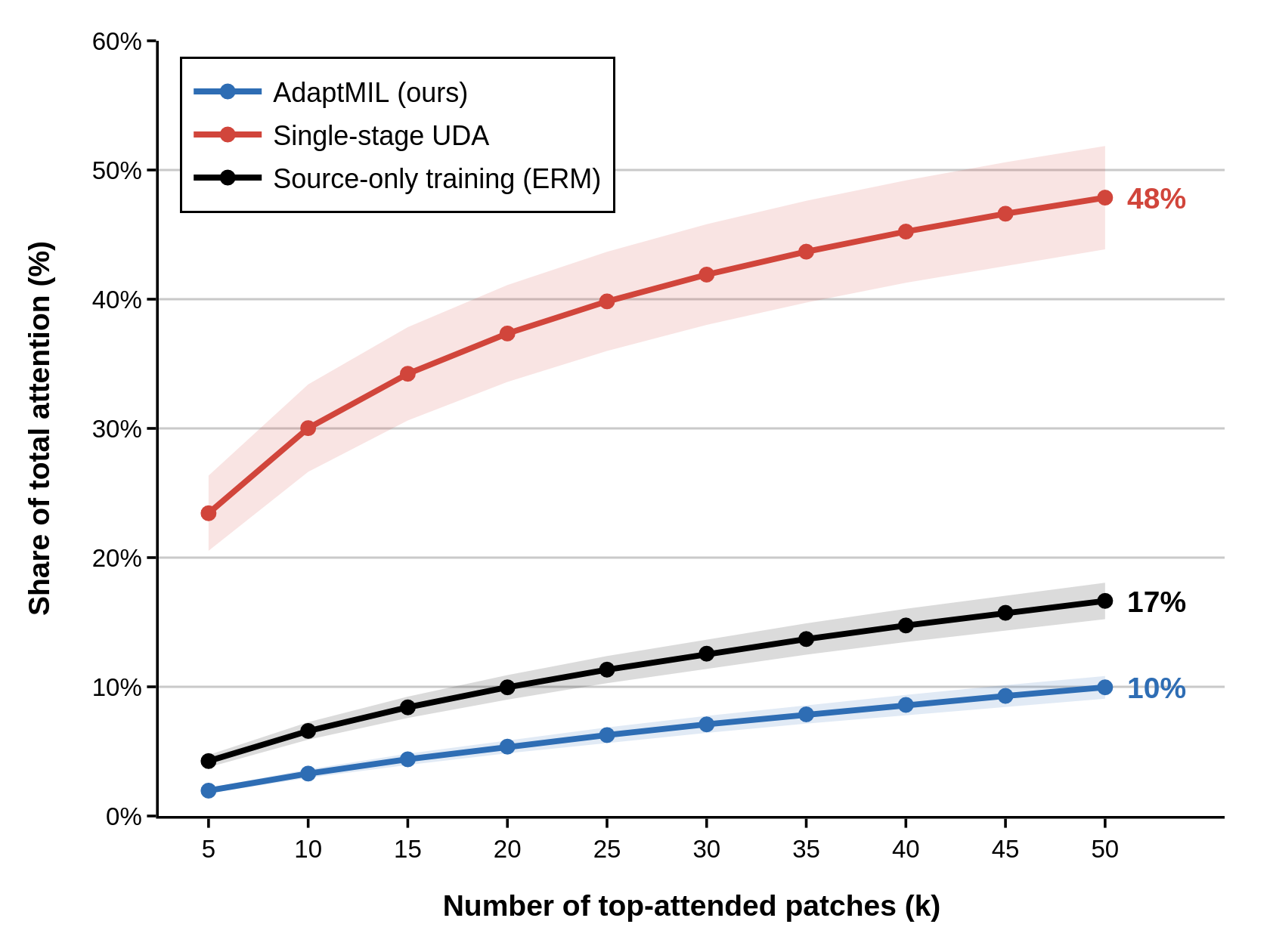}
        \caption{}
        \label{fig:analysis:attention}
    \end{subfigure}
    \hfill
    \begin{subfigure}[t]{0.48\linewidth}
        \centering
        \includegraphics[width=\linewidth]{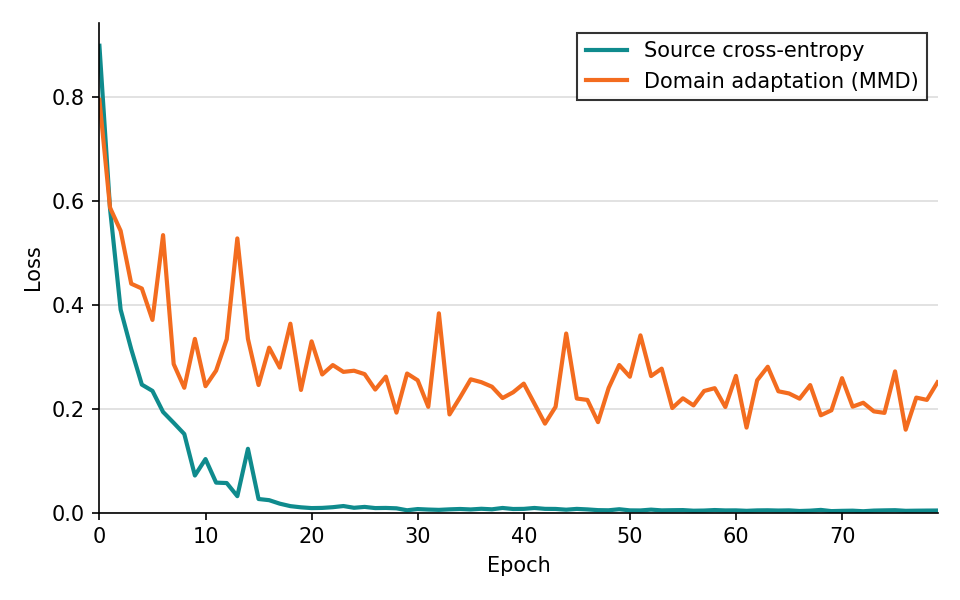}
        \caption{}
        \label{fig:analysis:loss}
    \end{subfigure}
    \caption{
    Analysis of the interaction between domain adaptation and ABMIL aggregation.
    (a) Standard UDA (single-stage) concentrates attention on a smaller subset of patches than source-only training.
    AdaptMIL reduces and even reverses this effect, resulting in a more distributed attention pattern.
    (b) In a representative single-stage UDA run, both source cross-entropy and MMD losses decrease despite degraded performance (the MMD plateau is expected).
    Results are from NSCLC subtyping with UNI; attention in (a) is averaged over held-out target slides.
    }
    \label{fig:analysis}
\end{figure}

In UDA, we are given $n_s$ labeled samples from a source domain $\{(x_i^s,y_i^s)\}_{i=1}^{n_s}$, together with $n_t$ unlabeled samples from a target domain, $\{x_j^t\}_{j=1}^{n_t}$.
In the cross-hospital setting, each sample $x$ is a WSI, the labels $y_i$ correspond to biological classes (e.g., cancer status), and each domain corresponds to a different hospital.
The goal is to learn a MIL module and a task head (e.g., a classifier) that generalize to the target hospital without having access to its labels.

Domain-invariant UDA methods 
jointly optimize a supervised task loss on labeled source samples and a domain adaptation objective that encourages alignment between source and target representations~\citep{wilson2020survey}.
Let $\vz_i^s = f_\theta(x_i^s)$ and $\vz_j^t = f_\theta(x_j^t)$ denote the source and target representations, respectively, and let $g_\phi$ be the task head.
The UDA training objective is
\begin{equation}
\label{eq:da_loss}
\frac{1}{n_s}\sum_{i=1}^{n_s}
\mathcal{L}_{\mathrm{task}}
\left(g_\phi(\vz_i^s), y_i^s\right)
+
\lambda\,
\mathcal{L}_{\mathrm{DA}}
\left(
\{\vz_i^s\}_{i=1}^{n_s},
\{\vz_j^t\}_{j=1}^{n_t}
\right),
\end{equation}
where $\mathcal{L}_{\mathrm{task}}$ is the supervised task loss, $\mathcal{L}_{\mathrm{DA}}$ is the domain adaptation loss, and $\lambda$ controls the tradeoff between them.
The task head and loss can correspond to any supervised prediction task, e.g., for classification, $g_\phi$ is a classifier and $\mathcal{L}_{\mathrm{task}}$ is the cross-entropy loss.

This setting can be adapted to slide-level MIL, where each sample is a bag of patches rather than a single image. 
Let $x_i^s=\{p_{i,1}^s,\ldots,p_{i,m_i^s}^s\}$ denote a source WSI decomposed into $m_i^s$ patches.
Each patch is encoded by a PFM encoder $f_{\mathrm{PFM}}$, yielding the bag of patch representations
$z_i^s=\{f_{\mathrm{PFM}}(p_{i,1}^s),\ldots,f_{\mathrm{PFM}}(p_{i,m_i^s}^s)\}$. 
The target WSI $x_j^t$ and bag $z_j^t$ are defined analogously.
A learnable MIL aggregation module $h_{\mathrm{MIL}}$ then maps $z_i^s$ into a slide representation, on which the task head operates.
Applying the UDA formulation gives
\begin{equation}
\label{eq:da_mil_loss}
\frac{1}{n_s}\sum_{i=1}^{n_s}
\mathcal{L}_{\mathrm{task}}
\left(
g_\phi(h_{\mathrm{MIL}}(z_i^s)), y_i^s
\right)
+
\lambda\,
\mathcal{L}_{\mathrm{DA}}
\left(
\{h_{\mathrm{MIL}}(z_i^s)\}_{i=1}^{n_s},
\{h_{\mathrm{MIL}}(z_j^t)\}_{j=1}^{n_t}
\right),
\end{equation}
where domain adaptation is performed on the slide representations produced by the MIL module.

\begin{figure}[t]
    \centering
    \begin{subfigure}{\linewidth}
        \centering
        \includegraphics[width=\linewidth]{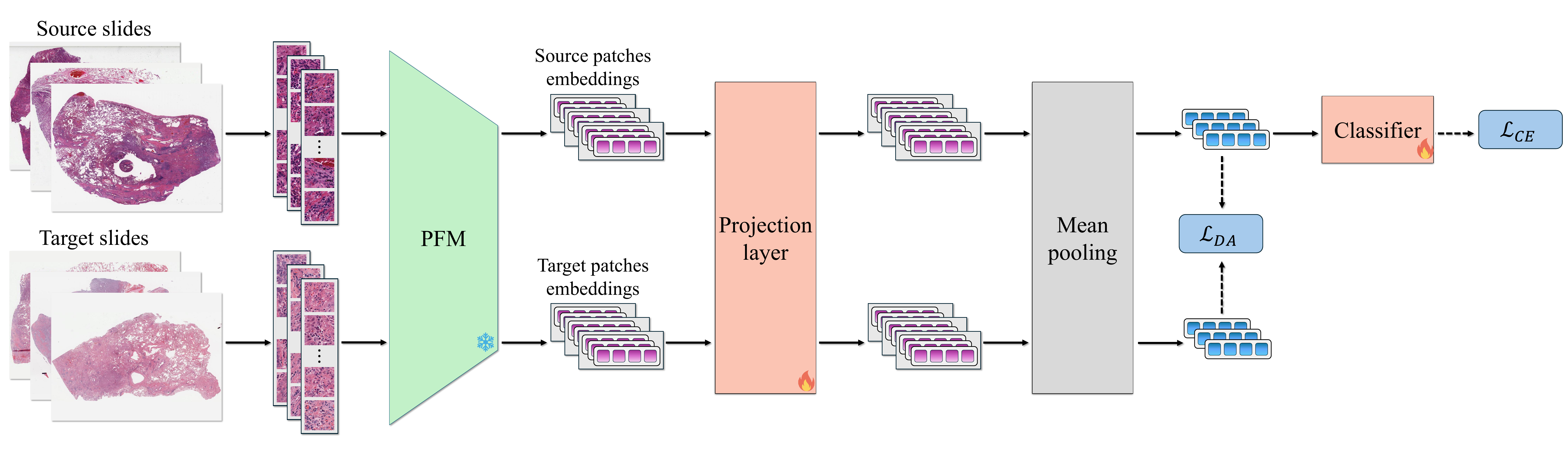}
        \caption{}
        \label{fig:top}
    \end{subfigure}
    \begin{subfigure}{\linewidth}
        \centering
        \includegraphics[width=\linewidth]{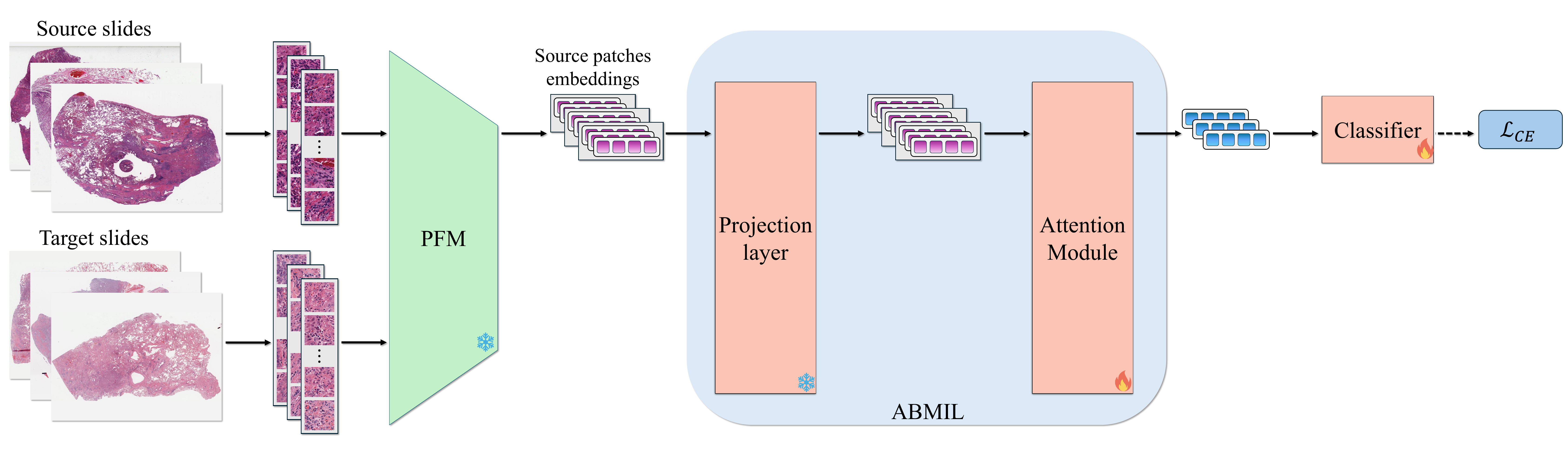}
        \caption{}
        \label{fig:bottom}
    \end{subfigure}
    \caption{
    AdaptMIL two-stage training with ABMIL for classification.
    (a) In Stage~1, a projection layer is trained to map patch embeddings into a more domain-aligned space.
    The projected embeddings are aggregated using mean pooling, and training uses both a cross-entropy (CE) loss on labeled source slides and a domain adaptation (DA) loss between source and target slide embeddings.
    (b) In Stage~2, mean pooling is replaced by ABMIL's learnable attention module. The projection layer is frozen, and the attention module and classifier are trained using only labeled source slides.}
    \label{fig:arch}
\end{figure}

As in standard domain-invariant UDA, the feature extractor (the PFM) can remain frozen while a learnable projection transforms its representations into a more domain-aligned space.
In many MIL architectures such a projection is already included prior to aggregation~\citep{shao2021transmil,ilse2018abmil,lu2021data}, eliminating the need for an additional projection layer.
The MIL module and task head are thus trained jointly,
with the only difference from standard source-only MIL training (using only labeled source data and the task loss) being the addition of the domain adaptation loss.

\subsection{Failure mode of domain adaptation and MIL training}
\label{subsec:method:failore_mode}

We find that the application of UDA to MIL training (\cref{eq:da_mil_loss}) is mostly ineffective, performing worse than source-only training in most settings and providing only limited gains otherwise (see~\cref{fig:adaptation_results}).
Analyzing this reveals a surprising effect: joint MIL and domain adaptation training substantially changes the learned attention pattern compared to standard MIL training, concentrating attention on a smaller subset of patches.
We demonstrate this effect in~\cref{fig:analysis:attention}:
Under UDA training, the top-$k$ attended patches in ABMIL account for a substantially larger fraction of the total attention (e.g., $2.8\times$ for $k=50$).
This shows that the adaptation objective affects not only the learned projection layer, as might be expected, but also how patch information is aggregated by the MIL module.

Examining the training dynamics shows that both the task and domain adaptation losses decrease as expected (\cref{fig:analysis:loss}), indicating that the optimization objectives are satisfied.
Together with the degradation in performance, this suggests that the MIL model takes advantage of a shortcut to reduce the domain gap.
That is, rather than learning genuinely more domain-robust representations, it can satisfy both objectives by changing how patch information is aggregated.

We hypothesize that this shortcut is facilitated by the morphological redundancy across patches in WSIs~\citep{song2024morphological}, which may provide the MIL model with multiple ways to satisfy the task objective.
Adding the domain adaptation constraint may therefore push the model toward a different aggregation solution that, in this case, generalizes worse.

We emphasize that more concentrated attention is not inherently undesirable, as its effect depends on the regions receiving high attention and the quality of the representations.
Our analysis therefore does not imply that concentrated attention alone causes performance degradation. 
Rather, it shows that domain adaptation can substantially alter the learned aggregation pattern, 
even though the intended effect is to only make representations more domain-invariant,
and that these changes are associated with degraded performance.

\subsection{AdaptMIL}
\label{subsec:method:adaptmil}

Our analysis shows that applying domain adaptation objectives interferes with MIL aggregation learning. 
This motivated us to try and decouple the two, namely separate domain-invariant representation learning from learning the aggregation mechanism.
We propose \textbf{AdaptMIL}, a two-stage training strategy that first learns domain-aligned representations and then learns the MIL aggregation over the adapted representations.
AdaptMIL is formulated here using ABMIL~\citep{ilse2018abmil} but can be applied to different MIL architectures. See~\cref{fig:arch} for an illustration of AdaptMIL.

ABMIL first transforms each patch representation through a learned projection layer $P$, 
and then aggregates the projected representations with an attention module $\mathrm{Attn}$.
Let $z_{i,k}=f_{\mathrm{PFM}}(p_{i,k})$ denote the representation of patch $k$ from slide $i$.
ABMIL is given by \(h_{\mathrm{MIL}}(z_i)=\mathrm{Attn}\!\left(P(z_i)\right)\), where \(P\) is applied independently to each patch representation, i.e., $P(z_i)=\{P(z_{i,1}),\ldots,P(z_{i,m_i})\}$.

In the first stage, AdaptMIL learns a projection layer $P$ that maps patch representations into a more domain-invariant space, and does not use the attention module.
To obtain slide-level representations for the task and domain alignment objectives without the attention module, the projected representations are aggregated using parameter-free mean pooling.
The resulting Stage~1 objective is:
\begin{equation}
\label{eq:adaptmil_stage1}
\begin{split}
\mathcal{L}_{\mathrm{stage1}}
&=
\frac{1}{n_s}\sum_{i=1}^{n_s}
\mathcal{L}_{\mathrm{task}}\!\left(
g_\phi\!\left(f_{\mathrm{mean}}(P(z_i^s))\right),\, y_i^s
\right) \\
&\quad + \lambda\, \mathcal{L}_{\mathrm{DA}}\!\left(
\{f_{\mathrm{mean}}(P(z_i^s))\}_{i=1}^{n_s},
\{f_{\mathrm{mean}}(P(z_j^t))\}_{j=1}^{n_t}
\right),
\end{split}
\end{equation}
where $f_{\mathrm{mean}}(P(z_i))=\frac{1}{m_i}\sum_{k=1}^{m_i}P(z_{i,k})$.
Mean pooling is a simple but competitive aggregation strategy in computational pathology~\citep{shao2025do}. Importantly, because it is parameter-free, it cannot be altered by the domain adaptation objective.

In the second stage, AdaptMIL learns the MIL aggregation over the adapted representations.
It keeps the projection layer $P$ learned in Stage~1 frozen and replaces mean pooling with ABMIL's attention module, recovering the full ABMIL architecture.
We train this stage using only source samples and the task loss. 
This allows the learnable attention module to focus on task-discriminative information without being influenced by the domain adaptation objective. The resulting Stage~2 objective is:
\begin{equation}
\label{eq:adaptmil_stage2}
\begin{split}
\mathcal{L}_{\mathrm{stage2}}
&=
\frac{1}{n_s}\sum_{i=1}^{n_s}
\mathcal{L}_{\mathrm{task}}\!\left(
g_\phi\!\left(h_{\mathrm{MIL}}(z_i^s)\right),\, y_i^s
\right).
\end{split}
\end{equation}

As shown in~\cref{fig:adaptation_results}, the decoupled two-stage training of AdaptMIL improves over both source-only training and standard UDA. \cref{fig:analysis:attention} shows that AdaptMIL does not exhibit the increased attention concentration observed with standard UDA and instead produces more distributed attention.
\section{Experiments}

\subsection{Tasks and datasets}

We evaluate AdaptMIL on four different tasks.
The first is non-small cell lung cancer (NSCLC) subtyping between lung adenocarcinoma and squamous cell carcinoma, using 489 WSIs from TCGA~\citep{komura22tcga-ut} across six hospitals, and 150 WSIs from an in-house hospital dataset.
The second task is renal cell carcinoma (RCC) subtyping between clear cell carcinoma and papillary cell carcinoma, using 307 WSIs from TCGA across three hospitals.
The third task is cancer detection, using 404 WSIs from CAMELYON17~\citep{bandi2018detection} across five hospitals.
The fourth task is survival prediction, presented in~\cref{supp:subsec:additional_results:survival} of the supplementary material (Supp.).

\subsection{Evaluation}

For each subtyping or detection dataset, we evaluate all source-target hospital pairs and report the average metric across them. For example, the TCGA NSCLC dataset contains six hospitals, resulting in 30 source-target pairs. For the in-house lung dataset, we evaluate both directions with each of the six TCGA hospitals, resulting in 12 pairs.
Each source-target pair is evaluated using 3-fold cross-validation, with evaluation performed on held-out target WSIs that are not used during training.
Our main metric for subtyping and detection is balanced accuracy.

\begin{table*}[!t]
    \centering
    \caption{
    Cross-hospital NSCLC TCGA subtyping results for AdaptMIL with ABMIL.
    The metric is balanced accuracy (\%). Values in parentheses denote the difference from ERM.}
    \begin{tabular}{lccc}
        \toprule
        \textbf{Method} &
        \textbf{UNI} &
        \textbf{CONCH-1.5} &
        \textbf{H0-mini} \\
        \midrule

        In-domain training
        & 86.98 {\dnone{$+7.21$}}
        & 88.88 {\dnone{$+2.93$}}
        & 91.38 {\dnone{$+7.07$}} \\
        
        \midrule
        Source-only (ERM)
        & 79.78
        & 85.95
        & 84.31 \\

        Macenko
        & 77.72 {\dneg{$-2.06$}}
        & 85.32 {\dneg{$-0.63$}}
        & 82.67 {\dneg{$-1.64$}} \\

        Patch-level fine-tuning
        & 84.25 {\dpos{$+4.47$}}
        & 87.11 {\dpos{$+1.16$}}
        & -- \\

        \midrule
        Single-stage UDA (MMD)
        & 82.22 {\dpos{$+2.45$}}
        & 86.75 {\dpos{$+0.80$}}
        & 80.91 {\dneg{$-3.40$}} \\
        AdaptMIL-MMD (ours)
        & \textbf{85.80} {\dpos{$+6.02$}}
        & \textbf{88.37} {\dpos{$+2.42$}}
        & 87.63 {\dpos{$+3.32$}} \\

        \addlinespace
        Single-stage UDA (LMMD)
        & 73.09 {\dneg{$-6.69$}}
        & 76.25 {\dneg{$-9.69$}}
        & 64.07 {\dneg{$-20.24$}} \\
        AdaptMIL-LMMD (ours)
        & 83.77 {\dpos{$+3.99$}}
        & 88.17 {\dpos{$+2.23$}}
        & 86.89 {\dpos{$+2.58$}} \\

        \addlinespace
        Single-stage UDA (CORAL)
        & 73.02 {\dneg{$-6.76$}}
        & 82.15 {\dneg{$-3.80$}}
        & 64.86 {\dneg{$-19.45$}} \\
        AdaptMIL-CORAL (ours)
        & 85.21 {\dpos{$+5.43$}}
        & 88.13 {\dpos{$+2.19$}}
        & \textbf{87.82} {\dpos{$+3.51$}} \\

        \addlinespace
        Single-stage UDA (DANN)
        & 79.37 {\dneg{$-0.41$}}
        & 85.17 {\dneg{$-0.78$}}
        & 84.49 {\dpos{$+0.18$}} \\
        AdaptMIL-DANN (ours)
        & 81.38 {\dpos{$+1.60$}}
        & 86.54 {\dpos{$+0.60$}}
        & 86.22 {\dpos{$+1.90$}} \\

        \bottomrule
    \end{tabular}
    \label{table:main_results}
\end{table*}

\subsection{Models and training}
We evaluate AdaptMIL across different PFMs, domain adaptation objectives, and MIL architectures.
For patch-level feature extraction, we use three PFMs: UNI~\citep{chen2024towards}, CONCH-1.5~\citep{lu2024visual}, and H0-mini~\citep{filiot2025distilling}. Tissue segmentation, patch extraction, and feature extraction are performed using the TRIDENT package~\citep{zhang2025standardizing}.

We evaluate four domain adaptation objectives: three discrepancy-based losses, MMD~\citep{long2015learning}, LMMD~\citep{zhu2020deep}, and CORAL~\citep{sun2016deep}, and the adversarial DANN objective~\citep{ganin2016domain}.
We further apply AdaptMIL to three MIL architectures: ABMIL~\citep{ilse2018abmil}, DSMIL~\citep{li2021dual}, and TITAN~\citep{ding2025multimodal}.
The implementation for ABMIL is described in~\cref{subsec:method:adaptmil}. The implementations for DSMIL and TITAN follow the same principle with architecture-specific adjustments.
All AdaptMIL models are trained for 80 epochs, with 40 epochs for each stage.
Further implementation details are provided in~\cref{supp:sec:implemenation} of the Supp.

\subsection{Method comparisons}

We compare AdaptMIL against several baselines. First, we compare against source-only training with standard MIL. This baseline, commonly referred to as empirical risk minimization (ERM)~\citep{vapnik1999overview}, represents the out-of-distribution performance that domain adaptation aims to improve.
We also compare to standard UDA applied to MIL (defined in~\cref{eq:da_mil_loss}), using each of the four domain adaptation objectives. We refer to this baseline as single-stage UDA.

We additionally compare against Macenko stain normalization~\citep{macenko2009method}. In this setting, the MIL model is trained only on source data, as in ERM, but evaluation is performed on target images whose appearance is normalized to a source reference. 
Finally, we compare against a domain adaptation approach that fine-tunes the PFM using patch-level supervision~\citep{vardi2026discrepancyminimizationimprovescrosshospital}.
We note that, to the best of our knowledge, other pathology-specific domain adaptation methods do not have publicly available implementations, and therefore we could not compare against them.

\begin{table*}[!t]
    \centering
    \caption{
    Cross-hospital classification results for AdaptMIL across different tasks.
    The metric is balanced accuracy (\%).
    All results use AdaptMIL based on ABMIL with MMD.
    IH-NS denotes the in-house NSCLC task, and Det. denotes cancer detection.
    }
    \setlength{\tabcolsep}{3.5pt}
    \begin{tabular}{lccccccccc}
        \toprule
        \multirow{2}{*}{\textbf{Method}} &
        \multicolumn{3}{c}{\textbf{UNI}} &
        \multicolumn{3}{c}{\textbf{CONCH-1.5}} &
        \multicolumn{3}{c}{\textbf{H0-mini}} \\
        \cmidrule(lr){2-4} \cmidrule(lr){5-7} \cmidrule(lr){8-10}
        &
        \textbf{IH-NS} & \textbf{Det.} & \textbf{RCC} &
        \textbf{IH-NS} & \textbf{Det.} & \textbf{RCC} &
        \textbf{IH-NS} & \textbf{Det.} & \textbf{RCC} \\

        \midrule

        \makecell[l]{In-domain training \\ {\scriptsize\phantom{($+0.00$)}}}
        & \makecell{89.49 \\ \dnone{$+8.87$}}
        & \makecell{82.25 \\ \dnone{$+9.43$}}
        & \makecell{96.45 \\ \dnone{$+4.67$}}
        & \makecell{89.44 \\ \dnone{$+5.70$}}
        & \makecell{98.41 \\ \dnone{$+1.99$}}
        & \makecell{93.79 \\ \dnone{$+0.79$}}
        & \makecell{93.02 \\ \dnone{$+12.03$}}
        & \makecell{84.93 \\ \dnone{$+3.75$}}
        & \makecell{94.94 \\ \dnone{$+2.10$}} \\

        \midrule

        Source-only (ERM)
        & 80.62
        & 72.82
        & 91.78
        & 83.74
        & \textbf{96.43}
        & 93.01
        & 81.00
        & 81.18
        & 92.83 \\

        \midrule

        \makecell[l]{Single-stage UDA \\ {\scriptsize\phantom{($+0.00$)}}}
        & \makecell{\textbf{85.20} \\ \dpos{$+4.59$}}
        & \makecell{72.31 \\ \dneg{$-0.51$}}
        & \makecell{90.75 \\ \dneg{$-1.03$}}
        & \makecell{85.58 \\ \dpos{$+1.84$}}
        & \makecell{86.11 \\ \dneg{$-10.31$}}
        & \makecell{92.53 \\ \dneg{$-0.47$}}
        & \makecell{76.44 \\ \dneg{$-4.56$}}
        & \makecell{70.28 \\ \dneg{$-10.90$}}
        & \makecell{88.93 \\ \dneg{$-3.90$}} \\

        \addlinespace[3pt]

        \makecell[l]{AdaptMIL (ours) \\ {\scriptsize\phantom{($+0.00$)}}}
        & \makecell{82.30 \\ \dpos{$+1.68$}}
        & \makecell{\textbf{78.87} \\ \dpos{$+6.05$}}
        & \makecell{\textbf{95.09} \\ \dpos{$+3.31$}}
        & \makecell{\textbf{86.72} \\ \dpos{$+2.98$}}
        & \makecell{95.59 \\ \dneg{$-0.84$}}
        & \makecell{\textbf{94.10} \\ \dpos{$+1.09$}}
        & \makecell{\textbf{86.92} \\ \dpos{$+5.92$}}
        & \makecell{\textbf{89.30} \\ \dpos{$+8.12$}}
        & \makecell{\textbf{93.32} \\ \dpos{$+0.48$}} \\

        \bottomrule
    \end{tabular}
    \label{table:datasets_results}
\end{table*}

\subsection{Results}

\cref{table:main_results} presents the evaluation of AdaptMIL with ABMIL across different PFMs and domain adaptation objectives on the NSCLC subtyping task.
AdaptMIL improves over ERM across all PFMs and objectives, with gains up to $+6.02\%$ in balanced accuracy.
MMD is the best-performing objective for two of the three PFMs, while LMMD and CORAL also yield notable gains, and DANN provides smaller improvements.
Similar trends are observed for ROC-AUC and F1 (see~\cref{table:main_results_additional} in Supp.).

We additionally report in-domain performance of an ABMIL model that is trained and evaluated on samples from the same hospital.
The performance drop from in-domain to source-only ERM highlights the degradation caused by the cross-hospital distribution shift.
AdaptMIL substantially reduces this gap. For example, with UNI and CONCH-1.5, AdaptMIL recovers more than $80\%$ of the gap.

\begin{table*}[!t]
    \centering
    \small
    \setlength{\tabcolsep}{2.5pt}

    \begin{minipage}[t]{0.47\linewidth}
        \centering
        \caption{
        Cross-hospital NSCLC TCGA subtyping results across different MIL architectures.
        The metric is balanced accuracy (\%).
        }
        \begin{tabular}{lccc}
            \toprule
            \textbf{Method} &
            \textbf{\makecell{ABMIL\\(UNI)}} &
            \textbf{\makecell{DSMIL\\(UNI)}} &
            \textbf{TITAN} \\
            \midrule

            In-domain training
            & \makecell{86.98\\\dnone{$+7.21$}}
            & \makecell{83.12\\\dnone{$+8.78$}}
            & \makecell{92.32\\\dnone{$+1.65$}} \\

            \midrule

            Source-only (ERM)
            & 79.78
            & 74.34
            & 90.68 \\

            \midrule

            Single-stage UDA
            & \makecell{82.22\\\dpos{$+2.45$}}
            & \makecell{81.29\\\dpos{$+6.95$}}
            & \makecell{89.50\\\dneg{$-1.18$}} \\

            AdaptMIL (ours)
            & \makecell{\textbf{85.80}\\\dpos{$+6.02$}}
            & \makecell{\textbf{82.10}\\\dpos{$+7.76$}}
            & \makecell{\textbf{90.88}\\\dpos{$+0.20$}} \\

            \bottomrule
        \end{tabular}
        \label{table:mil_results}
    \end{minipage}
    \hfill
    \begin{minipage}[t]{0.52\linewidth}
        \centering
        \caption{
        Ablation studies for AdaptMIL.
        }
        \begin{tabular}{lc}
            \toprule
            \textbf{Method} &
            \textbf{Balanced Acc.} \\
            \midrule

            Source-only (ERM)
            & 79.78 \\

            \midrule

            Single-stage + mean-pool adap. head
            & 84.87 \dnone{$+5.09$} \\

            AdaptMIL w/ Stage 1 only
            & 82.17 \dnone{$+2.39$} \\

            AdaptMIL w/ adaptation in Stage 2
            & 83.23 \dnone{$+3.45$} \\

            AdaptMIL w/o projection freeze
            & 85.56 \dnone{$+5.79$} \\

            AdaptMIL w/o adaptation loss
            & 81.38 \dnone{$+1.60$} \\

            \midrule

            AdaptMIL (full)
            & \textbf{85.80} \dnone{$+6.02$} \\

            \bottomrule
        \end{tabular}
        \label{table:ablations}
    \end{minipage}

\end{table*}
    
AdaptMIL outperforms the other adaptation methods.
As observed in~\cref{subsec:method:failore_mode}, directly applying standard single-stage UDA during MIL training often fails to improve over ERM and can substantially degrade performance. For example, LMMD with H0-mini reduces balanced accuracy by $20.24\%$.
Macenko normalization also fails to improve performance, consistent with previous reports~\citep{huang2025knowledge,lin2502unveiling}.
Patch-level fine-tuning of PFMs is the baseline that provides the largest gains over ERM, but it remains inferior to AdaptMIL.
Although effective, patch-level fine-tuning requires both patch supervision and re-extraction of patch features for every source--target adaptation setting.
In contrast, AdaptMIL requires only slide-level supervision and operates on pre-extracted PFM features, allowing the same features to be reused across adaptation settings.

We evaluate AdaptMIL across additional datasets and classification tasks in~\cref{table:datasets_results}.
AdaptMIL improves over source-only ERM in eight of the nine settings, with gains of up to $+8.12\%$.
The only exception is cancer detection with CONCH-1.5, where ERM already achieves very strong performance. 
In contrast, single-stage UDA degrades performance in seven of the nine settings.

Finally, we evaluate AdaptMIL with MIL architectures beyond ABMIL in~\cref{table:mil_results}.
AdaptMIL improves performance with DSMIL, although standard single-stage UDA is also effective. We further obtain a gain when adapting the pretrained TITAN slide encoder.
These results indicate that AdaptMIL is not specific to ABMIL and can be applied to other MIL architectures.

\subsection{Ablations}

We perform ablation studies to examine the impact of AdaptMIL's components, with results for AdaptMIL-MMD with ABMIL and UNI in~\cref{table:ablations}.
We first consider an alternative way to decouple representation alignment from aggregation learning.
Instead of two-stage training, we train ABMIL throughout with the task objective, while applying the adaptation objective to a separate mean-pooling head placed after the projection layer. 
This prevents adaptation gradients from reaching the attention module, while still adapting the projection layer.
This alternative method performs well (row~2 in~\cref{table:ablations}), but remains inferior to AdaptMIL, suggesting an additional benefit from explicitly separating alignment and aggregation learning into distinct training stages.

We next examine whether the second stage of AdaptMIL is necessary.
Specifically, we use only Stage~1, where the learned projection is followed by mean pooling and trained with both the task and adaptation objectives.
This performs worse than full AdaptMIL (row~3), confirming the benefit of using ABMIL, consistent with prior work favoring ABMIL over mean pooling~\citep{shao2025do}.

We additionally evaluate two design choices in Stage 2.
Keeping the adaptation loss active in Stage 2 degrades performance (row~4), supporting our hypothesis that adaptation gradients interfere with aggregation learning.
Not freezing the projection layer in Stage 2 slightly reduces performance (row~5), possibly because the projection becomes more source-oriented.

Finally, we test whether the two-stage training schedule itself contributes to the improvement by removing the adaptation objective (row~6).
This variant still improves over ERM, suggesting that separating representation and aggregation learning provides some benefit on its own.
However, it remains substantially below AdaptMIL, demonstrating the importance of the alignment in Stage 1.
\section{Conclusion}
\label{sec:conclusion}

In this paper, we introduce AdaptMIL, a training strategy for performing domain adaptation within MIL architectures. 
Motivated by the observation that applying standard UDA to MIL can interfere with aggregation learning and degrade performance, AdaptMIL decouples domain alignment and aggregation learning into two training stages.
Evaluated in cross-hospital adaptation settings with different tasks, PFMs, MIL architectures, and domain adaptation objectives, AdaptMIL consistently improves over source-only training and outperforms alternative adaptation methods.

Our work has several limitations.
We primarily evaluate MIL models trained from scratch,
which is the most common setting and offers flexibility in the choice of the PFM~\citep{shao2025do}.
Another approach is to use pretrained MIL models, which avoid training from scratch but are tied to specific PFM features.
Our TITAN experiments show that AdaptMIL can be applied in this setting, although broader evaluation remains for future work.
A further limitation is that AdaptMIL depends on the quality of the PFM representations and may not be able to compensate for strong representation biases.
Combining AdaptMIL with end-to-end PFM fine-tuning~\citep{kumar2026single} is therefore an interesting direction for future work.
Finally, our survival prediction results (in Supp.) show that AdaptMIL improves in this setting, while standard UDA can sometimes be effective, motivating further study of when decoupling domain alignment from MIL aggregation is necessary.

We focus on cross-hospital domain adaptation in digital pathology, but AdaptMIL may also apply to other MIL settings affected by a domain shift.
Examples include shifts between frozen and formalin-fixed paraffin-embedded tissues in pathology~\citep{ozyoruk2022deep}, as well as MIL applications in other domains such as medical imaging~\citep{cheplygina2019not} and remote sensing~\citep{paheding2024advancing}.

\bibliography{iclr2027_conference}
\bibliographystyle{iclr2027_conference}

\clearpage
{\LARGE\bfseries Supplementary Material\par}

\appendix
\section{Implementation details}
\label{supp:sec:implemenation}

We provide the implementation details used across the evaluated PFMs and MIL architectures.
We use the AdamW optimizer~\citep{loshchilov2018decoupled} with a cosine learning rate schedule~\citep{loshchilov2017sgdr}, starting at $10^{-4}$ and decaying to zero.
Each AdaptMIL stage uses its own cosine schedule (the schedule is restarted at the beginning of Stage~2).
The classifier is the only module trained in both stages and is carried over from Stage~1 to Stage~2.

We use a batch size of 16, with each batch balanced across source and target domains (i.e., 8 source samples and 8 target samples).
We additionally balance the labeled source samples within each batch by class, using an equal number of examples from each class.
The adaptation coefficient $\lambda$ is 2.5 for MMD, 1.0 for LMMD, 15 for CORAL, and 1.0 for DANN (also see~\cref{supp:subsec:additional_results:ablations}).

For the domain-adaptation objectives, we build on public implementations.
We use multi-kernel MMD, CORAL, and DANN from TLlib~\citep{tllib}, and LMMD from Transfer Learning Repo~\citep{transferlearning}.
All experiments are conducted in mixed precision on a single NVIDIA GeForce RTX 5090 GPU.

\paragraph{AdaptMIL for ABMIL.}

Our ABMIL implementation is based on MIL-Lab~\citep{shao2025do}.
The slide embedding dimension and internal attention dimension are set to match the dimensionality of each PFM's patch representations.

\paragraph{AdaptMIL for DSMIL.}
We use the multi-class DSMIL formulation~\citep{li2021dual}, with the implementation from MIL-Lab~\citep{shao2025do}.
Unlike ABMIL, DSMIL produces a class-specific slide representation for each class rather than a single slide embedding.
For domain adaptation, we concatenate the two class-specific representations and apply the adaptation objective to the resulting embedding.
The remaining implementation details follow ABMIL, as DSMIL similarly applies a patch-level projection followed by attention-based aggregation.

\paragraph{AdaptMIL for TITAN.}
TITAN~\citep{ding2025multimodal} consists of a pre-trained transformer that takes CONCH-1.5~\citep{lu2024visual} patch embeddings as input and produces contextualized patch representations,
followed by a pooling module that aggregates them into a single slide embedding.
A natural AdaptMIL decomposition would treat the transformer as the projection component and the pooling module as the aggregation component.
However, due to memory constraints and TITAN's limited batch-size support, we keep the transformer frozen and insert a trainable linear projection between the transformer and the pooler.
The projection is trained in Stage~1. In Stage~2, we keep it frozen and fine-tune the pooling module.

\paragraph{Baselines.}
We describe the implementation details of the different baselines.
For a fair comparison, single-stage UDA uses the same training configuration as AdaptMIL, except that training is performed in a single stage.
We train it for 80 epochs, matching the total number of epochs used across the two AdaptMIL stages.

For Macenko stain normalization~\citep{macenko2009method}, we use the implementation from torchstain~\citep{barbano2022torchstain}.
For each source hospital, we manually select a reference patch containing a substantial amount of tissue.

\section{Datasets}
\label{supp:sec:datasets}

We evaluate on four tasks across four public datasets: TCGA~\citep{komura22tcga-ut} NSCLC subtyping, TCGA RCC subtyping, CAMELYON17~\citep{bandi2018detection} cancer detection, and TCGA KIRC survival prediction.
\cref{tab:dataset_composition} summarizes the distribution of samples across hospitals for these datasets.
For CAMELYON17, we use only negative slides and slides containing macrometastases.
For the in-house NSCLC dataset, we use 150 slides, balanced between lung adenocarcinoma and lung squamous cell carcinoma. For all datasets, we have assured no patient leakage between train and test sets.

We note that some hospital datasets are relatively small and may exhibit substantial class imbalance which reflects a realistic cross-hospital setting.
Upon publication, we will release the exact slide lists used from all public datasets together with the code to facilitate reproducibility.

\begin{table*}[t]
    \centering
    \setlength{\tabcolsep}{2.5pt}
    \caption{
    Dataset composition for (a) TCGA NSCLC subtyping, (b) CAMELYON17 cancer detection, (c) TCGA RCC subtyping, and (d) TCGA KIRC survival prediction.
    LUAD: lung adenocarcinoma;
    LUSC: lung squamous cell carcinoma;
    ccRCC: clear cell renal cell carcinoma;
    pRCC: papillary renal cell carcinoma;
    IGC: International Genomics Consortium;
    RUMC: Radboud University Medical Center;
    CWZ: Canisius-Wilhelmina Hospital;
    UMCU: University Medical Center Utrecht;
    RST: Rijnstate Hospital;
    LPON: Laboratorium Pathologie Oost-Nederland;
    UNC: University of North Carolina;
    MSKCC: Memorial Sloan Kettering Cancer Center;
    UOP: University of Pittsburgh;
    MDA: MD Anderson.
    }
    \begin{subtable}[t]{0.48\textwidth}
    \centering
    \caption{TCGA NSCLC subtyping.}
    \label{tab:data_nsclc}
    \begin{tabular}{lrrr}
    \toprule
    \textbf{Hospital} & \textbf{LUAD} & \textbf{LUSC} & \textbf{Total} \\
    \midrule
    Asterand & 34 & 50 & 84 \\
    Christiana Healthcare & 47 & 20 & 67 \\
    Roswell Park & 12 & 25 & 37 \\
    Univ.\ of Pittsburgh & 22 & 13 & 35 \\
    Johns Hopkins & 84 & 50 & 134 \\
    IGC & 91 & 41 & 132 \\
    \midrule
    \textbf{Total} & 290 & 199 & 489 \\
    \bottomrule
    \end{tabular}
    \end{subtable}
    \hfill
    \begin{subtable}[t]{0.48\textwidth}
    \centering
    \caption{CAMELYON17 cancer detection.}
    \label{tab:data_camelyon}
    \begin{tabular}{lrrr}
    \toprule
    \textbf{Hospital} & \textbf{Negative} & \textbf{Macro} & \textbf{Total} \\
    \midrule
    RUMC & 64 & 15 & 79 \\
    CWZ & 58 & 12 & 70 \\
    UMCU & 75 & 14 & 89 \\
    RST & 60 & 19 & 79 \\
    LPON & 61 & 26 & 87 \\
    \midrule
    \textbf{Total} & 318 & 86 & 404 \\
    \bottomrule
    \end{tabular}
    \end{subtable}
    \vspace{3mm}
    
    \begin{subtable}[t]{0.48\textwidth}
    \centering
    \caption{TCGA RCC subtyping.}
    \label{tab:data_rcc}
    \begin{tabular}{lrrr}
    \toprule
    \textbf{Hospital} & \textbf{ccRCC} & \textbf{pRCC} & \textbf{Total} \\
    \midrule
    IGC & 53 & 31 & 84 \\
    UNC & 33 & 16 & 49 \\
    MSKCC & 139 & 35 & 174 \\
    \midrule
    \textbf{Total} & 225 & 82 & 307 \\
    \bottomrule
    \end{tabular}
    \end{subtable}
    \hfill
    \begin{subtable}[t]{0.48\textwidth}
    \centering
    \caption{TCGA KIRC survival prediction.}
    \label{tab:data_survival}
    \begin{tabular}{lrrr}
    \toprule
    \textbf{Hospital} &
    \textbf{Patients} &
    \textbf{Events} &
    \textbf{Censored} \\
    \midrule
    UOP & 104 & 69 & 35 \\
    MSKCC & 139 & 38 & 101 \\
    MDA & 69 & 26 & 43 \\
    \midrule
    \textbf{Total} & 312 & 133 & 179 \\
    \bottomrule
    \end{tabular}
    \end{subtable}

    \label{tab:dataset_composition}
\end{table*}

\section{Statistical tests}

We provide details on the statistical tests reported in~\cref{supp:sec:additional_results} for balanced accuracy and top-50 attention mass.
Statistical significance is assessed using a two-sided Wilcoxon signed-rank test over the 30 source--target hospital combinations in the TCGA NSCLC dataset.
For each source--target combination, we first average the relevant metric across the three cross-validation folds and then compute the difference between the compared methods (e.g., AdaptMIL vs.\ source-only ERM or single-stage UDA vs.\ source-only ERM).
We note that the Wilcoxon signed-rank test assumes independence across these differences, whereas in our setting the same hospital appears in multiple source--target combinations.
We therefore interpret the statistical results with this limitation in mind.

\section{Additional results}
\label{supp:sec:additional_results}

\subsection{Full NSCLC results}

We provide full results for the TCGA NSCLC subtyping experiments presented in \cref{table:main_results} of the main paper.
We report macro F1 and ROC-AUC results in~\cref{table:main_results_additional}.
The trends are consistent with the balanced accuracy results: AdaptMIL improves performance with MMD, LMMD, CORAL, and DANN.
The gains with DANN are generally smaller, consistent with known optimization challenges of gradient-reversal-based adversarial adaptation~\citep{tzeng2017adversarial}.

We additionally evaluate whether the balanced accuracy improvements of AdaptMIL over ERM are statistically significant, using a two-sided Wilcoxon signed-rank test with $\alpha=0.05$.
For MMD, LMMD, and CORAL, the improvements are statistically significant with UNI and H0-mini.
With CONCH-1.5, AdaptMIL improves the average performance, but the improvements are not statistically significant.
While these tests provide evidence of consistent gains across source--target combinations, AdaptMIL does not guarantee improvement for every individual combination.
For example, with MMD, AdaptMIL matches or outperforms ERM in 86.7\% of the combinations with UNI, but in only 63.3\% of the combinations with CONCH-1.5.

\begin{table*}[!t]
    \centering
    \caption{
    Cross-hospital TCGA NSCLC subtyping results for AdaptMIL with ABMIL, for (a) Macro F1 and (b) ROC-AUC. Metrics are reported on a 0--100 scale.
    }
    \begin{subtable}{\textwidth}
        \centering
        \subcaption{Macro F1.}
        \begin{tabular}{lccc}
            \toprule
            \textbf{Method} &
            \textbf{UNI} &
            \textbf{CONCH-1.5} &
            \textbf{H0-mini} \\
            \midrule

            In-domain training
            & 87.12 \dnone{$+10.71$}
            & 88.89 \dnone{$+5.05$}
            & 91.35 \dnone{$+8.51$} \\

            \midrule
            Source-only (ERM)
            & 76.41
            & 83.83
            & 82.84 \\

            Macenko
            & 72.85 \dneg{$-3.57$}
            & 83.22 \dneg{$-0.61$}
            & 79.77 \dneg{$-3.07$} \\

            \midrule
            Single-stage UDA (MMD)
            & 78.95 \dpos{$+2.54$}
            & 83.92 \dpos{$+0.08$}
            & 78.43 \dneg{$-4.41$} \\

            AdaptMIL-MMD (ours)
            & \textbf{83.54} \dpos{$+7.12$}
            & 86.35 \dpos{$+2.52$}
            & 86.27 \dpos{$+3.43$} \\

            \addlinespace
            Single-stage UDA (LMMD)
            & 69.36 \dneg{$-7.05$}
            & 73.13 \dneg{$-10.70$}
            & 60.67 \dneg{$-22.17$} \\

            AdaptMIL-LMMD (ours)
            & 80.86 \dpos{$+4.44$}
            & 86.33 \dpos{$+2.49$}
            & 85.81 \dpos{$+2.97$} \\

            \addlinespace
            Single-stage UDA (CORAL)
            & 70.29 \dneg{$-6.12$}
            & 79.49 \dneg{$-4.34$}
            & 61.10 \dneg{$-21.74$} \\

            AdaptMIL-CORAL (ours)
            & 83.14 \dpos{$+6.73$}
            & \textbf{86.42} \dpos{$+2.58$}
            & \textbf{86.53} \dpos{$+3.69$} \\

            \addlinespace
            Single-stage UDA (DANN)
            & 75.94 \dneg{$-0.47$}
            & 82.98 \dneg{$-0.85$}
            & 83.01 \dpos{$+0.17$} \\

            AdaptMIL-DANN (ours)
            & 78.62 \dpos{$+2.20$}
            & 84.92 \dpos{$+1.08$}
            & 85.24 \dpos{$+2.40$} \\

            \bottomrule
        \end{tabular}
        \label{table:main_results_f1}
    \end{subtable}
    
    \vspace{3mm}
    
    \begin{subtable}{\textwidth}
        \centering
        \subcaption{ROC-AUC.}
        \begin{tabular}{lccc}
            \toprule
            \textbf{Method} &
            \textbf{UNI} &
            \textbf{CONCH-1.5} &
            \textbf{H0-mini} \\
            \midrule

            In-domain training
            & 94.89 \dnone{$+3.09$}
            & 95.44 \dnone{$+0.31$}
            & 96.88 \dnone{$+3.30$} \\

            \midrule
            Source-only (ERM)
            & 91.80
            & 95.13
            & 93.57 \\

            Macenko
            & 91.94 \dpos{$+0.14$}
            & 94.82 \dneg{$-0.32$}
            & 93.53 \dneg{$-0.05$} \\

            \midrule
            Single-stage UDA (MMD)
            & 89.93 \dneg{$-1.87$}
            & 95.21 \dpos{$+0.08$}
            & 88.12 \dneg{$-5.46$} \\

            AdaptMIL-MMD (ours)
            & \textbf{94.38} \dpos{$+2.59$}
            & 96.29 \dpos{$+1.16$}
            & \textbf{95.05} \dpos{$+1.47$} \\

            \addlinespace
            Single-stage UDA (LMMD)
            & 81.26 \dneg{$-10.54$}
            & 84.19 \dneg{$-10.94$}
            & 69.88 \dneg{$-23.69$} \\

            AdaptMIL-LMMD (ours)
            & 93.93 \dpos{$+2.13$}
            & 95.90 \dpos{$+0.77$}
            & 94.95 \dpos{$+1.37$} \\

            \addlinespace
            Single-stage UDA (CORAL)
            & 80.27 \dneg{$-11.52$}
            & 89.67 \dneg{$-5.46$}
            & 71.75 \dneg{$-21.82$} \\

            AdaptMIL-CORAL (ours)
            & 93.96 \dpos{$+2.16$}
            & \textbf{96.41} \dpos{$+1.27$}
            & 94.74 \dpos{$+1.16$} \\

            \addlinespace
            Single-stage UDA (DANN)
            & 91.65 \dneg{$-0.15$}
            & 94.29 \dneg{$-0.84$}
            & 93.67 \dpos{$+0.10$} \\

            AdaptMIL-DANN (ours)
            & 94.12 \dpos{$+2.32$}
            & 96.31 \dpos{$+1.18$}
            & 94.65 \dpos{$+1.08$} \\

            \bottomrule
        \end{tabular}
        \label{table:main_results_auc}
    \end{subtable}
    \label{table:main_results_additional}
\end{table*}

\subsection{Additional attention and optimization analysis}

To further support the attention analysis in~\cref{subsec:method:failore_mode} and~\cref{fig:analysis:attention}, we perform two-sided Wilcoxon signed-rank tests ($\alpha=0.05$) on the top-50 attention mass in the TCGA NSCLC dataset.
Across all 12 PFM and adaptation method combinations, single-stage UDA yields a higher average top-50 attention mass than ERM on held-out target slides, with the difference being statistically significant in 11 of the 12 settings.

Conversely, AdaptMIL produces a more distributed attention pattern than ERM in all 12 settings, with statistically significant differences in 11 of the 12 settings.
We hypothesize that this occurs because Stage~1 learns the projection using mean pooling, which aggregates patch representations uniformly, and this affects the attention patterns subsequently learned in Stage~2.

\begin{figure}[t]
    \centering
    \includegraphics[width=0.7\linewidth]{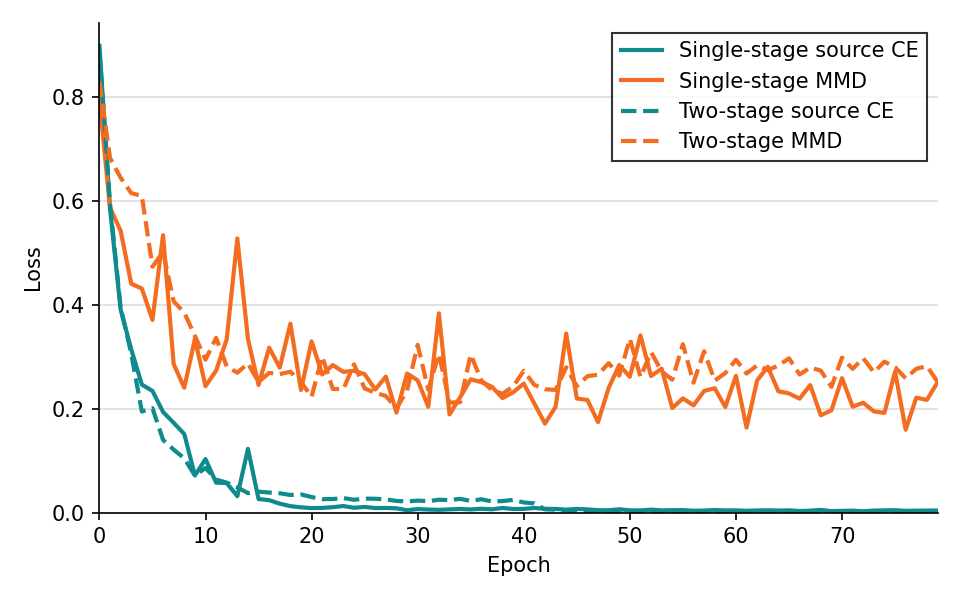}
    \caption{
    Representative training dynamics of single-stage UDA and AdaptMIL, showing similar behavior of the source cross-entropy and MMD losses during training.
    For AdaptMIL, the MMD loss is not optimized during Stage~2 (after epoch 40) and is shown only for monitoring.
    }
    \label{fig:supp_losses}
\end{figure}

Lastly, for completeness, we show that the source cross-entropy and domain adaptation losses exhibit similar dynamics in single-stage UDA and AdaptMIL.
\cref{fig:supp_losses} extends \cref{fig:analysis:loss} by also showing the AdaptMIL losses, which follow a similar pattern.
Despite the similar optimization behavior, single-stage UDA performs $3.21\%$ below ERM in this example, whereas AdaptMIL improves over ERM by $17.65\%$, further supporting our hypothesis that single-stage UDA exploits a shortcut.

\subsection{Survival prediction}
\label{supp:subsec:additional_results:survival}

In the main paper, we evaluate AdaptMIL on three classification tasks. 
Here, we additionally consider a non-classification task: survival prediction, which ranks patients according to their disease-specific survival risk.
We use 312 patients from the TCGA kidney renal clear cell carcinoma (KIRC) cohort across three hospitals.
Evaluation is performed using leave-one-out hospital evaluation with 5-fold cross-validation, and performance is measured using the concordance index (C-Index).

Implementation uses the same ABMIL-based training framework, replacing the classification head and cross-entropy loss with a single-output risk head that predicts each patient's log relative hazard. The model is trained with the negative Cox partial log-likelihood.

\cref{table:survival_results} presents the survival prediction results.
AdaptMIL improves over ERM for all evaluated PFMs.
However, unlike the classification tasks, single-stage UDA is also effective and even outperforms AdaptMIL, although the margins are mostly small.

We observe an increase in attention concentration under single-stage UDA, as shown in~\cref{fig:analysis:attention} for NSCLC subtyping and in the other classification tasks.
However, the increase in top-50 attention mass is much smaller for survival prediction. For example, with UNI and MMD it increases by only $1.2\times$, compared with $2.8\times$ for NSCLC subtyping.
This suggests that the interaction between domain adaptation and MIL aggregation may depend on the downstream task. Nevertheless, AdaptMIL still improves over source-only training across all evaluated PFMs, suggesting a more robust choice when the effectiveness of single-stage UDA is uncertain.

\begin{table}[!t]
    \centering
    \caption{
        Cross-hospital survival prediction results for AdaptMIL-MMD with ABMIL.
        The metric is C-index.
    }
    \begin{tabular}{lccc}
        \toprule
        \textbf{Method} &
        \textbf{UNI} &
        \textbf{\makecell{CONCH-1.5}} &
        \textbf{\makecell{H0-mini}} \\
        \midrule

        In-domain training
        & \makecell{0.6303\\\dnone{$+0.0173$}}
        & \makecell{0.6185\\\dnone{$-0.0162$}}
        & \makecell{0.6352\\\dnone{$+0.0105$}} \\

        \midrule

        Source-only (ERM)
        & 0.6130
        & 0.6346
        & 0.6247 \\

        \midrule

        Single-stage UDA
        & \makecell{\textbf{0.6406}\\\dpos{$+0.0276$}}
        & \makecell{\textbf{0.6494}\\\dpos{$+0.0147$}}
        & \makecell{\textbf{0.6520}\\\dpos{$+0.0272$}} \\

        AdaptMIL (ours)
        & \makecell{0.6372\\\dpos{$+0.0242$}}
        & \makecell{0.6490\\\dpos{$+0.0144$}}
        & \makecell{0.6377\\\dpos{$+0.0130$}} \\

        \bottomrule
    \end{tabular}
    \label{table:survival_results}
\end{table}

\subsection{Additional ablations}
\label{supp:subsec:additional_results:ablations}

\begin{table*}[t]
    \centering
      \caption{
        Sensitivity of AdaptMIL to the domain adaptation coefficient $\lambda$
        on NSCLC subtyping with UNI and ABMIL.
    }

    \setlength{\tabcolsep}{5pt}

    \begin{subtable}[t]{0.46\textwidth}
        \centering
        \caption{MMD.}
        \begin{tabular}{lccc}
            \toprule
            \textbf{Objective} &
            \textbf{$\lambda$} &
            \textbf{Balanced Acc.} &
            \textbf{$\Delta$} \\
            \midrule

            \multirow{13}{*}{MMD}
            & 0.5  & 84.84 & +5.07 \\
            & 1.0  & 85.22 & +5.44 \\
            & 1.5  & 85.48 & +5.70 \\
            & 2.0  & 85.67 & +5.89 \\
            & 2.5  & \textbf{85.80} & \textbf{+6.02} \\
            & 3.0  & \textbf{85.80} & \textbf{+6.02} \\
            & 3.5  & 85.57 & +5.79 \\
            & 4.0  & 85.74 & +5.96 \\
            & 4.5  & 85.70 & +5.92 \\
            & 5.0  & 85.64 & +5.87 \\
            & 10.0 & 85.46 & +5.68 \\
            & 15.0 & 85.41 & +5.63 \\
            & 20.0 & 85.42 & +5.64 \\

            \bottomrule
        \end{tabular}
        \label{tab:lambda_sensitivity_mmd}
    \end{subtable}
    \hfill
    \begin{subtable}[t]{0.50\textwidth}
        \centering
        \caption{LMMD and CORAL.}
        \begin{tabular}{lccc}
            \toprule
            \textbf{Objective} &
            \textbf{$\lambda$} &
            \textbf{Balanced Acc.} &
            \textbf{$\Delta$} \\
            \midrule

            \multirow{5}{*}{LMMD}
            & 0.5 & 83.09 & +3.31 \\
            & 1.0 & 83.77 & +3.99 \\
            & 1.5 & \textbf{84.16} & \textbf{+4.39} \\
            & 2.0 & 83.77 & +4.00 \\
            & 2.5 & 83.21 & +3.43 \\

            \midrule

            \multirow{7}{*}{CORAL}
            & 1.0  & 82.94 & +3.16 \\
            & 1.5  & 83.66 & +3.88 \\
            & 2.5  & 83.94 & +4.16 \\
            & 5.0  & 84.74 & +4.96 \\
            & 10.0 & 85.13 & +5.36 \\
            & 15.0 & \textbf{85.21} & \textbf{+5.43} \\
            & 20.0 & 85.14 & +5.36 \\

            \bottomrule
        \end{tabular}
        \label{tab:lambda_sensitivity_lmmd_coral}
    \end{subtable}

     \label{tab:lambda_sensitivity}
\end{table*}

We evaluate the sensitivity of AdaptMIL to the domain adaptation coefficient $\lambda$ in~\cref{tab:lambda_sensitivity}.
AdaptMIL is relatively insensitive to the choice of $\lambda$ across MMD, LMMD, and CORAL.
For example, across different MMD coefficients, the difference between the best and worst balanced accuracy is only $0.96\%$.

\subsection{Runtime}
We measure the training time of AdaptMIL-MMD with UNI.
Training a single source--target model takes 27.6 seconds on average for a TCGA NSCLC fold, compared with 47.6 seconds for single-stage UDA and 21.7 seconds for source-only ERM.
AdaptMIL is therefore approximately $1.7\times$ faster than single-stage UDA.
This is because Stage~1 uses the computationally cheaper mean-pooling aggregation, and Stage~2 does not compute the domain adaptation objective. In contrast, single-stage UDA uses both learned MIL aggregation and the adaptation objective throughout training.

\end{document}